\documentclass[conference]{IEEEtran}
\IEEEoverridecommandlockouts
\usepackage{amsmath,amssymb,amsfonts}
\usepackage{graphicx}
\usepackage{textcomp}
\usepackage{xcolor}
\usepackage{enumitem}
\def\BibTeX{{\rm B\kern-.05em{\sc i\kern-.025em b}\kern-.08em
    T\kern-.1667em\lower.7ex\hbox{E}\kern-.125emX}}

\makeatletter % changes the catcode of @ to 11
\newcommand{\linebreakand}{%
  \end{@IEEEauthorhalign}
  \hfill\mbox{}\par
  \mbox{}\hfill\begin{@IEEEauthorhalign}
}
\makeatother % changes the catcode of @ back to 12

\usepackage{booktabs}
\usepackage{multirow}
\usepackage[labelfont=bf,textfont=it]{caption}
\usepackage{subcaption}
\usepackage[sort&compress,numbers]{natbib}
\usepackage{listings}
\usepackage{eso-pic} % arXiv only: copyright notice

\usepackage{hyperref}
\hypersetup{
    colorlinks,
    linkcolor={blue},
    citecolor={blue},
    urlcolor={blue},
    breaklinks=true,
    plainpages=true
}

\newcommand{\cref}[3]{\hyperref[#2]{#1~\ref*{#2}{#3}}}
\newcommand{\colref}[2]{\hyperref[#2]{#1~\ref*{#2}}}

\newcommand{\figref}[1]{\colref{Fig.}{#1}}

\newcommand{\secref}[1]{\colref{Section}{#1}}

\newcommand{\tabref}[1]{\colref{Table}{#1}}

\begin{document}
\IEEEoverridecommandlockouts
% arXiv: the camera-ready \IEEEpubid text is blanked (space kept so pagination matches ICLAD) and the IEEE
% accepted-manuscript notice is stamped in the bottom margin of page 1.
\IEEEpubid{\makebox[\columnwidth]{ }\hspace{\columnsep}\makebox[\columnwidth]{ }}
\AddToShipoutPictureBG*{\AtPageLowerLeft{\put(\LenToUnit{0.5\paperwidth},\LenToUnit{0.3in}){\makebox[0pt][c]{\parbox[b]{\textwidth}{\scriptsize\copyright~2026 IEEE. Personal use of this material is permitted. Permission from IEEE must be obtained for all other uses, in any current or future media, including reprinting/republishing this material for advertising or promotional purposes, creating new collective works, for resale or redistribution to servers or lists, or reuse of any copyrighted component of this work in other works. Accepted for publication at the 2026 IEEE International Conference on LLM-Aided Design (ICLAD).}}}}}

% \title{CRAFT: A Multi-Agent Pipeline for Zero-Shot\\Text-to-Parametric CAD Generation}
\title{CRAFT: Corrective and Robust Multi-Agent Framework for Text-to-Parametric CAD\\
% \thanks{a. Iowa State University, USA}
% \thanks{b. New York University, USA}
% \thanks{This work was supported by the National Science Foundation under grant CMMI-AM-2347623/2347624. We would like to thank the NVIDIA Corporation for providing GPUs for academic research.}
}

% Double-blind: no author info

%\author{\IEEEauthorblockN{Anonymous Authors}
%\IEEEauthorblockA{Anonymous Institution}}

% Musthafa, Anushrut, Mahdi, Chinmay, Aditya, Adarsh
\author{\IEEEauthorblockN{Mohammed Musthafa Rafi\textsuperscript{a}
\and Anushrut Jignasu\textsuperscript{a}
\and Mahdi Saraeian\textsuperscript{a}
\and Chinmay Hegde\textsuperscript{b}
\and Aditya Balu\textsuperscript{a}
\linebreakand
Adarsh Krishnamurthy\textsuperscript{a*}}
\IEEEauthorblockA{\textsuperscript{a}Iowa State University, USA \qquad \textsuperscript{b}New York University, USA}
\IEEEauthorblockA{Email: \textsuperscript{a} (mohd7 | anushrutg | mehdi | baditya | adarsh)@iastate.edu , \textsuperscript{b} chinmay.h@nyu.edu}
}

\maketitle

\begin{abstract}
Text-to-CAD generation has the potential to make mechanical CAD more accessible, but existing approaches face a trade-off between data dependence and output fidelity. Supervised methods rely on large paired text--model datasets, while zero-shot prompting strategies are often brittle and tend to lose the parametric structure needed for downstream editing and customization. We present CRAFT, a corrective and robust multi-agent framework that produces parametric OpenSCAD programs from natural language without task-specific training. At its core is a JSON intermediate representation that carries symbolic parametric expressions throughout the generation process, so the resulting models remain editable and retain their parametric structure. CRAFT combines semantic understanding, parametric planning, compilation, rendering, self-correction, and component-level verification within a layered recovery process, using multi-view visual feedback to detect and repair geometric errors. Across NopSCADlib, ABC, and Slice-100K, CRAFT is competitive with direct zero-shot baselines in perceptual and geometric quality while exposing substantially more editable parametric structure.
\end{abstract}


\section{Introduction}
\label{Sec:intro}

Text-driven geometric modeling has seen rapid progress in recent years, with large language and vision models enabling increasingly capable systems for generating 3D shapes directly from natural language descriptions~\citep{zhang2026llmcad,poole2023dreamfusion,jun2023shape,nichol2022pointe}. These advances have raised the prospect of making mechanical computer-aided design (CAD) accessible to users without specialized CAD expertise, reducing the barrier between a design intent expressed in words and a manufacturable geometric model. Despite this progress, existing approaches remain brittle in practice. Supervised methods that fine-tune models on paired text--CAD datasets can produce geometrically plausible outputs within their training distribution. They generalize poorly to novel shapes, however, and require substantial curated data to train~\citep{text2cad2024,cadcoder2025,text2cadquery2025,li2025llm4cad}. Zero-shot prompting strategies avoid this data dependency but are prone to generating syntactically invalid code, non-renderable geometry, and outputs that fail to faithfully reflect the original prompt. Both families of approaches also tend to produce geometry with fixed numeric dimensions rather than editable parametric models, limiting their utility for downstream engineering and customization tasks~\citep{wu2021deepcad,fusion360gallery2021}.

Agentic systems offer a natural path toward addressing these limitations~\citep{guo2024multiagent,madaan2023selfrefine}. Rather than treating CAD generation as a single forward pass, a multi-agent pipeline can decompose the problem into discrete stages, including semantic understanding, structured planning, code synthesis, and verification, each of which can be independently validated and repaired. This staged architecture allows errors to be detected and corrected locally rather than propagating silently into the final output, improving both reliability and geometric fidelity. A particularly well-suited target for such a pipeline is OpenSCAD~\citep{openscad}, a declarative scripting language in which geometry is expressed as a program over symbolic parameters. Generating OpenSCAD code rather than mesh geometry preserves the parametric relationships that make CAD models editable and reusable. However, verifying the correctness of generated OpenSCAD programs is difficult: a program may compile successfully yet produce geometry that is visually or dimensionally incorrect relative to the prompt. Effective verification therefore requires grounding outputs in external reference knowledge, both to check that recognized components conform to standard specifications and to confirm that expected parts are present. NopSCADlib~\citep{nopscadlib}, a parametric OpenSCAD component library, provides catalog-defined specifications for standard mechanical parts, grounding outputs in physically meaningful dimensions and enabling verification of component completeness.

\begin{figure*}[t!]
    \centering
    \includegraphics[width=0.75\linewidth, trim=20 10 40 20, clip]{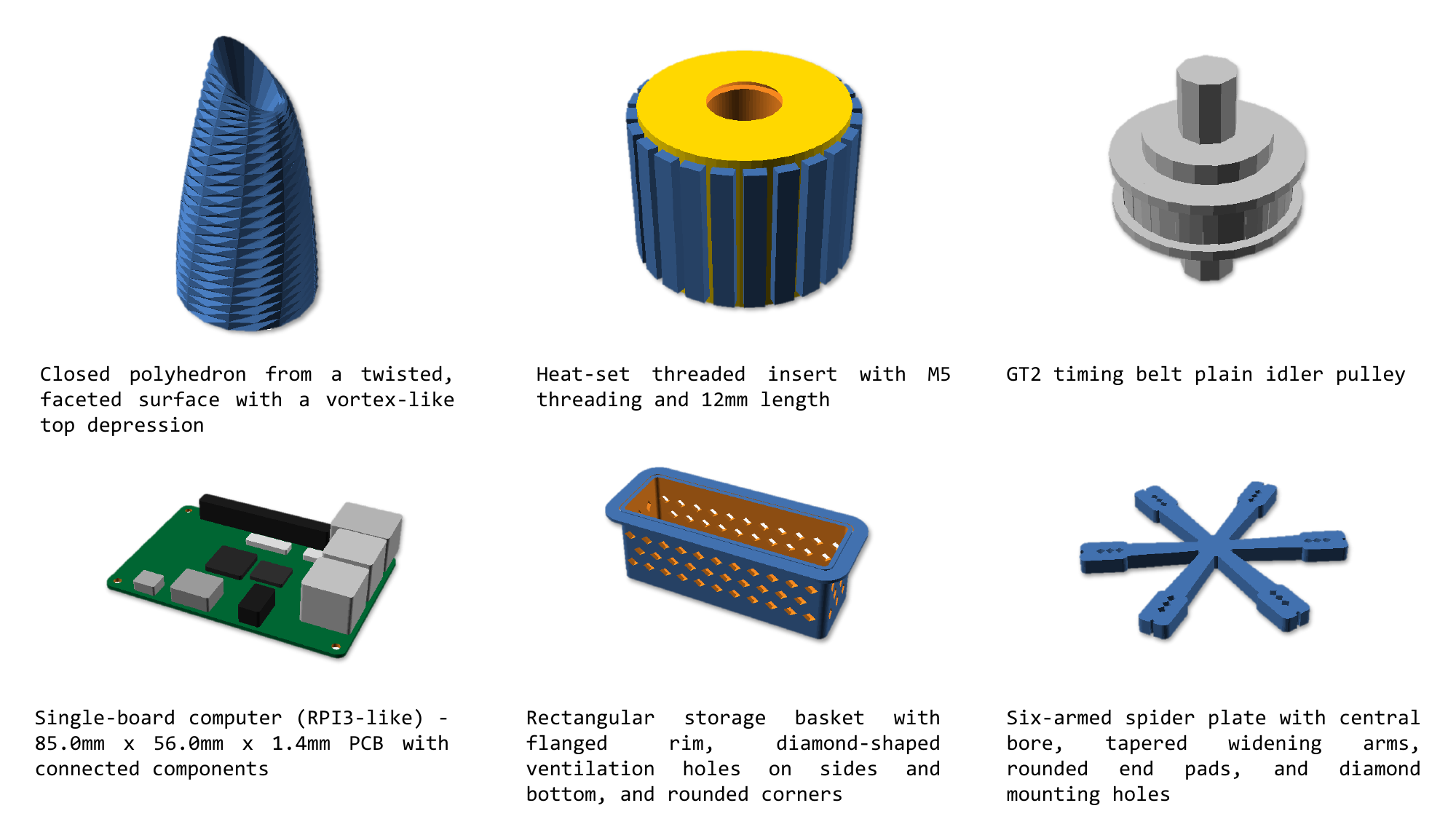}
    \caption{Examples of different CAD models created by CRAFT from the prompts.}
    \label{Fig:teaser}
\end{figure*}

We present CRAFT (\figref{Fig:teaser}), a corrective and robust multi-agent framework for text-to-parametric CAD generation. CRAFT produces executable and editable OpenSCAD programs from natural language without task-specific training through three core mechanisms: a JSON-based intermediate representation that preserves symbolic parametric expressions, multi-view visual feedback to detect and correct geometric errors, and a layered recovery strategy to handle failures at every stage of the pipeline. Our main contributions are as follows:
\begin{itemize}[topsep=0pt]
\item We develop a corrective and robust six-stage multi-agent framework for zero-shot text-to-parametric CAD generation that combines structured planning, multi-view visual verification, and layered error recovery to produce executable OpenSCAD programs from natural language.
\item We propose a JSON-based intermediate representation that preserves symbolic parametric expressions throughout the generation process, enabling the produced CAD programs to remain editable and reusable rather than hardening into fixed numeric geometry.
\item We evaluate CRAFT across three datasets (NopSCADlib, ABC, and Slice-100K) and show that it achieves competitive perceptual and geometric quality and substantially greater editability relative to direct generation baselines, and we isolate the contribution of each framework component through an ablation study.
\end{itemize}

Our implementation, generated OpenSCAD outputs, and evaluation scripts are publicly available at \url{https://github.com/idealab-isu/CRAFT}.

\section{Related Work}
\label{Sec:related}

\textbf{Text-to-3D and Text-to-CAD Generation.} Generative approaches to 3D modeling have advanced rapidly, spanning neural implicit representations~\citep{poole2023dreamfusion}, point cloud synthesis~\citep{nichol2022pointe}, and shape generation from text~\citep{jun2023shape}. While these methods produce visually compelling outputs, they typically yield fixed mesh geometry that cannot be edited parametrically or reused in engineering workflows. A parallel line of work targets CAD-specific representations: Text2CAD's sequential construction sequences~\citep{text2cad2024}, LLM4CAD's multimodal generation~\citep{li2025llm4cad}, CAD-Coder's chain-of-thought fine-tuning with geometric rewards~\citep{cadcoder2025}, and Text-to-CadQuery's scaled paired-data training~\citep{text2cadquery2025}. These supervised approaches achieve strong in-distribution results but require curated paired training data and produce outputs that are often not directly editable as parametric programs. CRAFT avoids task-specific training entirely and instead uses structured zero-shot generation to produce symbolic, editable OpenSCAD programs.

\textbf{Program Synthesis and Code Generation.} Generating executable code from natural language has become a well-studied problem, with large language models demonstrating strong performance across programming tasks~\citep{chen2021codex}. Techniques such as chain-of-thought prompting~\citep{wei2022chainofthought} and self-refinement~\citep{madaan2023selfrefine} have shown that decomposing generation into reasoning steps improves correctness on structured outputs, motivating intermediate representations and iterative repair for CAD, where generated programs must satisfy syntactic validity, successful compilation, and geometric and semantic correctness. CRAFT operationalizes this through a JSON intermediate representation that decouples planning from code synthesis, enabling schema-level validation before any OpenSCAD code is produced.

\begin{figure*}[t]
    \centering
    \includegraphics[width=0.95\linewidth]{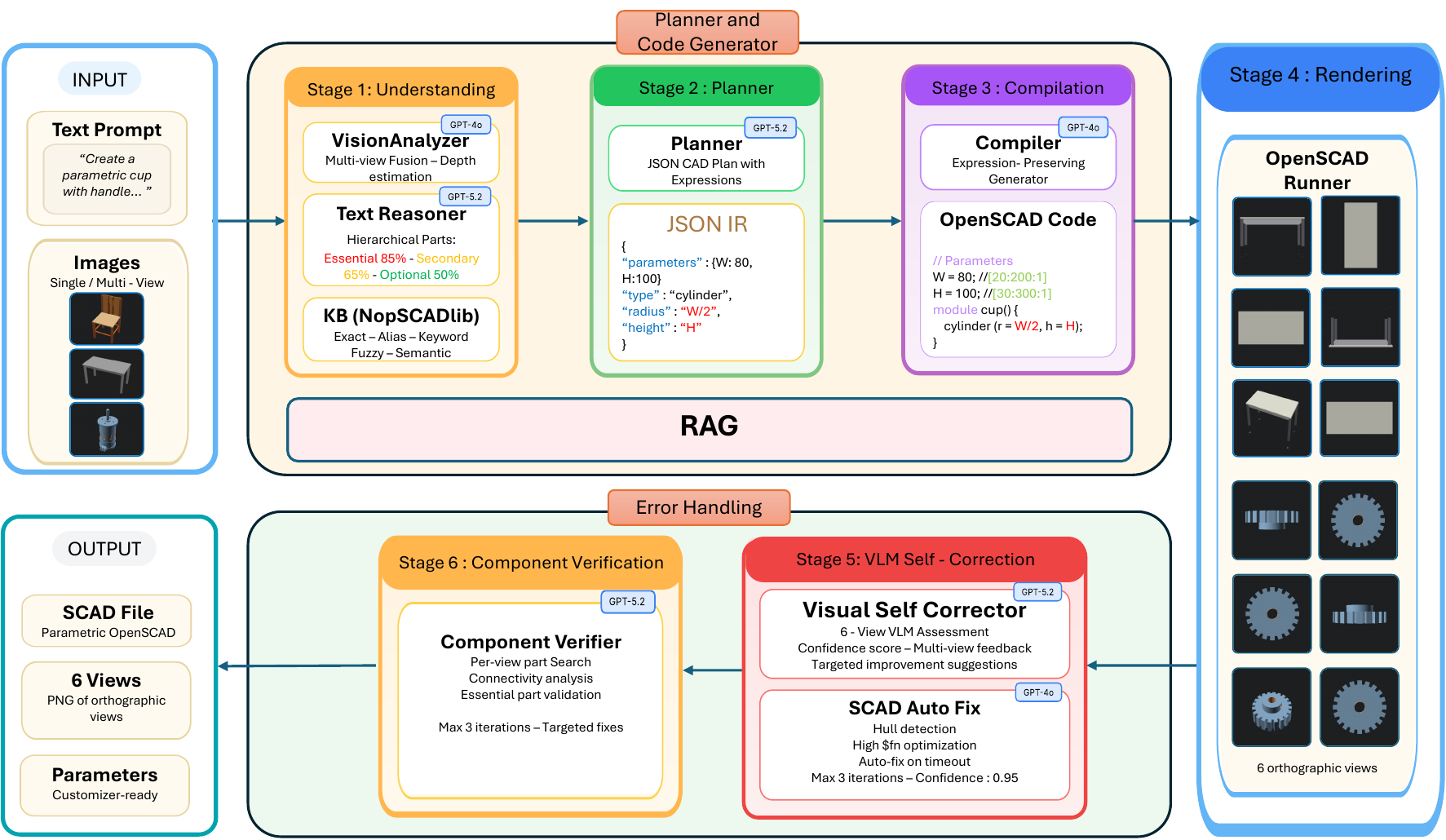}
    \caption{CRAFT maps a text prompt and optional reference images to a parametric OpenSCAD program through six stages: understanding, planning, compilation, rendering, vision-language model (VLM) self-correction, and component verification. Reasoning-intensive stages use GPT-5.2, while deterministic translation and repair steps use GPT-4o.}
    \label{Fig:pipeline}
\end{figure*}

\textbf{Verification and Visual Feedback.} Recent work addresses correctness verification as a first-class concern in CAD generation. CADCodeVerify~\citep{alrashedy2024cadcodeverify} introduces vision-language model-based verification and iterative correction as a post-generation step, improving quality without retraining. More broadly, visual feedback has emerged as a powerful signal for program repair, providing information difficult to obtain from code analysis alone. CRAFT extends this by rendering six orthographic views, enabling the VLM to detect errors not visible from any single view, and integrating this within a bounded correction loop instead of a post-processing step.

\textbf{Retrieval-Augmented Generation.} Retrieval-augmented generation (RAG) grounds language model outputs in external knowledge, improving factual accuracy and domain specificity~\citep{lewis2020rag}, and in code generation reduces hallucination by retrieving relevant library components at generation time. CRAFT applies this to parametric CAD by retrieving component specifications from NopSCADlib~\citep{nopscadlib} during the understanding stage, grounding geometry in catalog-defined dimensions rather than the model's implicit knowledge.

\textbf{Multi-Agent Systems and Parametric Modeling.} Multi-agent pipelines outperform single-model generation on tasks requiring planning, execution, and verification across multiple steps~\citep{guo2024multiagent}, isolating failure modes and applying targeted repair without regenerating earlier stages. This suits CAD generation, where failures arise at the planning, compilation, rendering, or geometric-correctness levels, each requiring a different repair strategy. The target representation matters as much as the architecture: parametric CAD expresses geometry through symbolic variables and editable relationships rather than fixed coordinates, enabling downstream customization, reuse, and analysis~\citep{fusion360gallery2021,wu2021deepcad}. Most text-to-CAD systems emit hard-coded numeric geometry; CRAFT instead stores dimensions as symbolic strings in its JSON intermediate representation and carries them through compilation unchanged, so the final OpenSCAD~\citep{openscad} programs remain fully editable.

\section{Methods}
\label{Sec:method}

CRAFT transforms natural language descriptions and optional reference images into parametric OpenSCAD code through six sequential stages. In our implementation, reasoning-intensive stages (text understanding, planning, multi-view assessment, and component verification) are handled by a reasoning large language model (LLM, GPT-5.2)~\citep{openai2025gpt52}, while image analysis and deterministic translation and repair steps are handled by a traditional LLM (GPT-4o)~\citep{openai2024gpt4o}. The six stages describe the forward generation path, whereas the five recovery layers introduced in \secref{subSec:recovery} describe how the system escalates repair when intermediate artifacts are invalid or the rendered result is unsatisfactory.

The framework (\figref{Fig:pipeline}) begins with understanding (Stage~1), where image and text inputs are processed in parallel. The image-analysis path extracts geometric cues from any provided reference images and fuses information across views when multiple images are available. The text-reasoning parses the prompt to identify required components, classify them by importance (essential, secondary, optional), and detect references to standard parts such as motors or fasteners. When such components are recognized, their specifications are retrieved from NopSCADlib~\citep{nopscadlib} using five strategies (exact name matching, curated alias lookup, keyword-based lookup, fuzzy string matching, and semantic embedding retrieval), following a retrieve-then-generate design pattern~\citep{lewis2020rag}. If retrieval fails, the framework proceeds with LLM-only synthesis.

The output of stage~1 is passed to planning (Stage~2), where the planner produces a JSON CAD plan (\secref{subSec:json-ir}). This intermediate representation encodes geometry, constructive solid geometry (CSG) operations, and symbolic parametric expressions in a structured format that can be validated before any code is generated. In compilation (Stage~3), the validated plan is translated into OpenSCAD that preserves expression strings and generates parameter ranges for the customizer interface.

In rendering (Stage~4), OpenSCAD produces six orthographic views (front, back, left, right, top, bottom), applying timeout-aware simplifications for expensive constructs such as \texttt{hull()}. These views drive VLM self-correction (Stage~5, \secref{subSec:vlm}). Finally, component verification (Stage~6) checks whether expected parts are present using tiered confidence thresholds of 85\% for essential parts, 65\% for secondary, and 50\% for optional. These thresholds were set empirically during pilot runs to prioritize functionally necessary components over cosmetic details, a process related to recent self-refinement approaches for large models~\citep{madaan2023selfrefine}.

\subsection{JSON Intermediate Representation}
\label{subSec:json-ir}

CRAFT uses a JSON intermediate representation (IR) between design understanding and OpenSCAD generation. The IR encodes parameters, geometric primitives, and CSG operations in a structured form that can be validated before compilation. The IR preserves symbolic parametric expressions and defers their evaluation: expressions such as \texttt{"outer\_diameter/2"} pass unchanged into the compiled OpenSCAD code, allowing the final model to remain editable and fully parametric.

\subsection{VLM Self-Correction}
\label{subSec:vlm}

Single-image assessment can miss geometric issues visible only from specific viewpoints: a model that looks correct from the front may have wrong depth or features visible only from the side. CRAFT addresses this by rendering six orthographic views and presenting them jointly to the VLM for assessment. The VLM returns a confidence score on a 0--1 scale, a list of identified issues, and suggested corrections. If the confidence falls below 0.95, the system generates targeted fixes for the affected parts rather than regenerating the entire design. This loop runs for up to three iterations, with early stopping once the confidence threshold is reached. In our implementation, this assessment is performed using GPT-5.2 due to its stronger spatial reasoning capability.

\subsection{Layered Error Recovery}
\label{subSec:recovery}

The recovery layers describe how CRAFT escalates repair when intermediate artifacts fail validation, rendering, or semantic checks. Failures in text-to-CAD generation can arise at several levels: invalid intermediate plans, non-renderable code, and geometrically incorrect outputs. CRAFT addresses these through a five-layer strategy that orders repair mechanisms from inexpensive structural fixes to more costly semantic correction. The first layer repairs malformed JSON plans through schema validation and automatic repair. The second layer applies deterministic OpenSCAD fixes for common rendering failures, such as expensive operations or excessive mesh resolution. The third layer re-invokes the multi-view VLM assessment from Stage~5 to detect semantic and geometric mismatches and generate targeted corrections. The fourth layer re-invokes the component verification procedure from Stage~6 under the same tiered thresholds. The fifth layer incorporates manual repair hints when automatic recovery remains insufficient. Automatic recovery operates under a fixed retry budget of 10 total attempts across the first four layers, set empirically to balance correction capability against latency and cost. In the reported experiments, staged recovery resolved compilation and rendering failures before final output assessment, allowing the comparison to focus on geometric fidelity and semantic alignment of the final recovered outputs.

\section{Experimental Setup}
\label{Sec:setup}

\textbf{Datasets.} We evaluate CRAFT on three datasets totaling 668 components to assess both in-domain performance and cross-dataset generalization. The primary benchmark is NopSCADlib~\citep{nopscadlib}, comprising 468 text--CAD pairs derived from a parametric OpenSCAD mechanical component library, with ground-truth geometry from catalog-defined parametric specifications. To test generalization beyond the retrieval library, we also evaluate on 100 shapes each from ABC~\citep{koch2019abc} and Slice-100K~\citep{slice100k2024}; these lie outside NopSCADlib and are run with retrieval disabled. Within each dataset, components are split into Simple, Medium, and Complex terciles by a weighted composite of structural features (NopSCADlib source code) or mesh-geometry features (external meshes); because the feature spaces differ, tiers are relative within a dataset rather than comparable across datasets.

\textbf{Baselines and configuration.} We compare against direct generation with GPT-4o~\citep{openai2024gpt4o} and GPT-5.2~\citep{openai2025gpt52}, which receive the same prompt and a system instruction to emit only valid OpenSCAD code in a single deterministic (temperature~$0.0$) call, with no intermediate planning, retrieval, visual feedback, or recovery. This evaluates CRAFT against standard one-shot direct generation from the same model family, although the methods use different inference budgets. CRAFT uses a hybrid allocation (GPT-5.2 for reasoning-intensive stages and GPT-4o for deterministic compilation and repair), with up to 2 schema-repair attempts, up to 3 VLM self-correction iterations (early stopping at confidence~$>0.95$), and up to 3 verification iterations under the tiered thresholds of \secref{Sec:method}.

\textbf{Metrics and protocol.} For geometric evaluation we report bidirectional Chamfer Distance (CD), F1 at a 1\% distance threshold, and voxel Intersection-over-Union (IoU) on a $64^3$ grid, computed from point clouds of 10{,}000 points sampled per mesh:
\begin{equation}
\operatorname{CD}=\tfrac{1}{2}\bigl(\operatorname{mean}(d_{\mathrm{gt}\to\mathrm{pred}})+\operatorname{mean}(d_{\mathrm{pred}\to\mathrm{gt}})\bigr).
\end{equation}
For perceptual evaluation we report CLIP score (ViT-L/14~\citep{radford2021clip}) for text--image alignment and Fr\'echet Inception Distance (FID~\citep{heusel2017fid}) from Inception-v3 features for distributional similarity to ground-truth renders. All views are rendered at $512\times512$ resolution. When a method fails to produce a valid output, the example is assigned worst-case metric values so that invalid generations remain reflected in the aggregates.

\section{Results}
\label{Sec:results}

We evaluate CRAFT against direct generation baselines (GPT-4o and GPT-5.2) across semantic alignment, geometric faithfulness, and editability, and ablate each component to isolate its contribution, spanning the in-domain NopSCADlib benchmark and the two external datasets of \secref{Sec:setup}.

\subsection{Perceptual Alignment}

\tabref{tab:perceptual} reports perceptual alignment. CRAFT attains the highest overall CLIP score, closest to the ground-truth reference, and the lowest overall FID. On FID, the margin is most pronounced on medium and complex components, while on simple parts GPT-5.2 attains the lowest FID, consistent with direct generation being effective for basic geometry. On CLIP score, CRAFT leads on simple and complex components, with GPT-5.2 narrowly ahead on medium. The Overall FID is lower than any per-tier value because it is pooled over all 468 renders rather than averaged across tiers; FID is non-additive, so the pooled estimate need not lie between tier values and carries less finite-sample bias.

\begin{table}[t!]
    \centering
    \caption{Perceptual evaluation on NopSCADlib. CLIP Score measures text--image alignment (higher is better), while FID measures distributional similarity to ground-truth renders (lower is better). GT denotes the ground-truth CLIP score as a reference value.}
    \label{tab:perceptual}
    \small
    \setlength{\tabcolsep}{4pt}
    \begin{tabular}{lcccc}
    \toprule
    Metric & CRAFT & GPT-4o & GPT-5.2 & GT \\
    \midrule
    CLIP Score & \textbf{0.2226} & 0.2033 & 0.2145 & 0.2342 \\
    \quad Simple & \textbf{0.2272} & 0.2098 & 0.2174 & 0.2279 \\
    \quad Medium & 0.2132 & 0.2014 & \textbf{0.2144} & 0.2342 \\
    \quad Complex & \textbf{0.2279} & 0.1983 & 0.2116 & 0.2409 \\
    \midrule
    FID & \textbf{95.11} & 119.42 & 106.22 & --- \\
    \quad Simple & 152.37 & 144.38 & \textbf{136.22} & --- \\
    \quad Medium & \textbf{126.88} & 159.56 & 147.95 & --- \\
    \quad Complex & \textbf{144.85} & 184.39 & 160.51 & --- \\
    \bottomrule
    \end{tabular}
\end{table}

\subsection{Geometric Accuracy}

We next evaluate geometric fidelity using Chamfer Distance, F1 (at a 1\% threshold), and voxel IoU. \tabref{tab:nopscad-results} reports in-domain results on NopSCADlib and \tabref{tab:external-results} the out-of-domain results on ABC and Slice-100K. On NopSCADlib, CRAFT attains the best overall Chamfer distance and the best overall voxel IoU and is comparable to the strongest baseline on F1. Direct generation remains competitive on the simpler in-distribution parts, whereas CRAFT's advantage is clearest on volumetric overlap and on the more complex components. We therefore characterize CRAFT's in-domain geometry as competitive and tied-or-better rather than dominant. On the external datasets, where retrieval is disabled, CRAFT generalizes well: it is best on every metric on ABC and best on Chamfer distance and F1 on Slice-100K, with GPT-4o edging voxel IoU on the latter. Taken together with the perceptual results, these findings indicate that CRAFT produces geometrically coherent parametric models with balanced quality across metrics. These numbers reflect end-to-end quality after the layered recovery of \secref{subSec:recovery}, whose per-stage behavior we quantify in \secref{subsec:robustness}.

\begin{table}[t!]
    \centering
    \caption{Geometric evaluation on NopSCADlib, including overall results and breakdown by prompt complexity. F1 is reported at a 1\% distance threshold. Best results in each row group are shown in bold.}
    \label{tab:nopscad-results}
    \small
    \setlength{\tabcolsep}{6pt}
    \begin{tabular}{llccc}
    \toprule
    Split & Method & CD$\downarrow$ & F1$\uparrow$ & Voxel IoU$\uparrow$ \\
    \midrule
    \multirow{3}{*}{Overall}
    & CRAFT   & \textbf{0.0654} & 0.2530 & \textbf{0.2356} \\
    & GPT-4o  & 0.0733 & 0.2467 & 0.2135 \\
    & GPT-5.2 & 0.0662 & \textbf{0.2587} & 0.1890 \\
    \midrule
    \multirow{3}{*}{Simple}
    & CRAFT   & 0.0805 & \textbf{0.3771} & 0.2863 \\
    & GPT-4o  & 0.0841 & 0.3399 & \textbf{0.3339} \\
    & GPT-5.2 & \textbf{0.0800} & 0.3475 & 0.1554 \\
    \midrule
    \multirow{3}{*}{Medium}
    & CRAFT   & 0.0594 & 0.2277 & 0.2431 \\
    & GPT-4o  & 0.0652 & 0.2262 & 0.1810 \\
    & GPT-5.2 & \textbf{0.0571} & \textbf{0.2357} & \textbf{0.2591} \\
    \midrule
    \multirow{3}{*}{Complex}
    & CRAFT   & \textbf{0.0559} & 0.1492 & \textbf{0.1744} \\
    & GPT-4o  & 0.0706 & 0.1694 & 0.1201 \\
    & GPT-5.2 & 0.0614 & \textbf{0.1883} & 0.1487 \\
    \bottomrule
    \end{tabular}
\end{table}

\begin{table}[t!]
\centering
\caption{Geometric evaluation on the external datasets (ABC and Slice-100K), with retrieval disabled. F1 is reported at a 1\% distance threshold. Best per dataset in bold.}
\label{tab:external-results}
\small
\setlength{\tabcolsep}{6pt}
\begin{tabular}{llccc}
\toprule
Dataset & Method & CD$\downarrow$ & F1$\uparrow$ & Voxel IoU$\uparrow$ \\
\midrule
\multirow{3}{*}{ABC}
& CRAFT   & \textbf{0.0841} & \textbf{0.1233} & \textbf{0.0433} \\
& GPT-4o  & 0.1008 & 0.0857 & 0.0212 \\
& GPT-5.2 & 0.0969 & 0.0875 & 0.0192 \\
\midrule
\multirow{3}{*}{Slice-100K}
& CRAFT   & \textbf{0.0612} & \textbf{0.2924} & 0.0616 \\
& GPT-4o  & 0.0751 & 0.2122 & \textbf{0.0711} \\
& GPT-5.2 & 0.0622 & 0.2768 & 0.0636 \\
\bottomrule
\end{tabular}
\end{table}

\subsection{Ablation Study}
\label{subsec:ablation}

% \tabref{tab:ablation}
We disable each framework component in turn on the full NopSCADlib benchmark to understand the effect of the removal of each component relative to the baseline (full framework). Removing retrieval ($-$retrieval) produces the largest degradation on voxel IoU, confirming the value of catalog-grounded component specifications for in-domain generation. Removing layered recovery ($-$recovery) produces the largest degradation on Chamfer distance and F1, and removing VLM self-correction ($-$multi-view) produces a comparable drop in F1 with a smaller increase in Chamfer distance. Removing the JSON intermediate representation ($-$JSON-IR) yields a similar Chamfer distance but lower F1 and voxel IoU; its primary role is to preserve the symbolic structure underlying CRAFT's editability (\secref{subsec:edit}) rather than to improve raw geometry. Removing component verification ($-$verification) slightly improves the mesh-based metrics, indicating that these metrics do not directly capture the semantic-completeness objective of the verification stage, which adds or repairs prompt-specified parts.

% \begin{table}[t]
% \centering
% \caption{Ablation on NopSCADlib: full framework versus single-component-disabled variants ($-$component). The final column summarizes the effect of removing each component relative to the full system.}
% \label{tab:ablation}
% \footnotesize
% \setlength{\tabcolsep}{3pt}
% \begin{tabular}{lcccl}
% \toprule
% Variant & CD$\downarrow$ & F1@1\%$\uparrow$ & IoU$\uparrow$ & Effect of removal \\
% \midrule
% CRAFT (full)      & 0.0654 & 0.2530 & 0.2356 & reference \\
% $-$\,retrieval    & 0.0764 & 0.2030 & 0.1545 & CD$\uparrow$ F1$\downarrow$ IoU$\downarrow$ \\
% $-$\,multi-view   & 0.0738 & 0.1907 & 0.2272 & CD$\uparrow$ F1$\downarrow$ \\
% $-$\,recovery     & 0.0773 & 0.1878 & 0.2289 & CD$\uparrow$ F1$\downarrow$ \\
% $-$\,JSON-IR      & 0.0643 & 0.2383 & 0.1999 & CD$\approx$ F1$\downarrow$ IoU$\downarrow$ \\
% $-$\,verification & 0.0637 & 0.2574 & 0.2411 & all$\uparrow$ slightly \\
% \bottomrule
% \end{tabular}
% \end{table}

\subsection{Editability}
\label{subsec:edit}

\begin{figure}[t]
    \centering
    \includegraphics[width=0.95\linewidth]{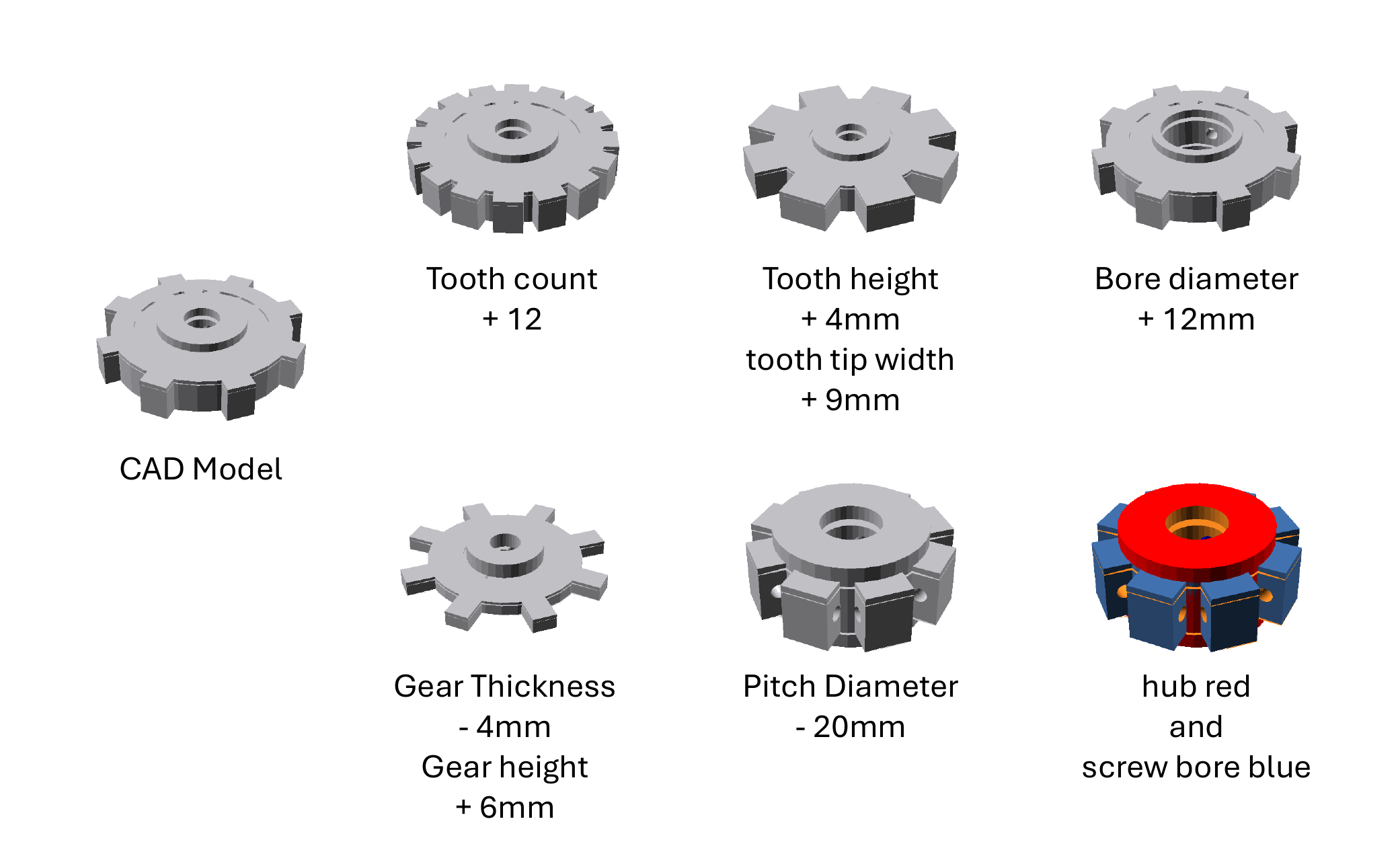}
    \caption{Preserved parametric editability on a generated spur gear: single- and multi-parameter edits produce coherent geometric and appearance changes. The bottom-right panel labels the hub (red) and screw bores (blue) for reference.}
    \label{Fig:parametric_edits}
\end{figure}

A central goal of CRAFT is to produce editable parametric programs rather than fixed geometry. We quantify this by perturbing the exposed top-level parameters of each generated program by $\pm10\%$ and $\pm50\%$ and re-rendering. The \emph{symbolic-preservation rate} is the fraction of secondary assignments whose value is an expression referencing other parameters rather than a hard-coded literal; \emph{edit success} is the fraction of (parameter, perturbation) attempts that apply, and \emph{post-edit validity} is the fraction whose perturbed program still renders to a valid, non-empty mesh. \tabref{tab:edit} shows that CRAFT exposes 13.1 editable parameters per model on average, against 4.2 for GPT-5.2 and 0.1 for GPT-4o, and preserves symbolic expressions in 66.8\% of cases versus 47.7\% and 3.0\%. Edit success and post-edit validity are uniformly high across methods, so the methods differ chiefly in how much editable structure they expose; edits apply cleanly regardless. The advantage widens with design difficulty: CRAFT's symbolic-preservation rate rises from 59.2\% on simple to 78.1\% on complex parts, with mean exposed parameters increasing from 10.2 to 17.6. \figref{Fig:parametric_edits} shows representative single- and multi-parameter edits on a generated spur gear.

\begin{table}[t]
\centering
\caption{Editability on NopSCADlib: exposed parameter count, symbolic-preservation rate, edit-success rate, and post-edit render validity.}
\label{tab:edit}
\small
\setlength{\tabcolsep}{5pt}
\begin{tabular}{lcccc}
\toprule
Method & \# Params & Symbolic$\uparrow$ & Edit succ.$\uparrow$ & Valid$\uparrow$ \\
\midrule
CRAFT   & \textbf{13.1} & \textbf{66.8\%} & 99.9\% & 99.1\% \\
GPT-5.2 & 4.2  & 47.7\% & 99.2\% & 97.7\% \\
GPT-4o  & 0.1  & 3.0\%  & 95.0\% & 95.0\% \\
\bottomrule
\end{tabular}
\end{table}

\subsection{Recovery Behavior}
\label{subsec:robustness}

\tabref{tab:recovery} reports per-stage recovery on NopSCADlib. The two inexpensive structural layers (schema repair, SCAD auto-fix) are rarely triggered but always succeed when invoked. VLM self-correction is the most frequently used layer, addressing 80.4\% of samples and resolving about half of those. Component verification is triggered on 17.3\% of samples and applies part-level corrections; because its objective is component completeness rather than the rendered-image confidence used to score the other layers, its recovery rate under that shared success flag is 0 (\tabref{tab:recovery}, note~\textsuperscript{*}). The manual-repair layer was never invoked, confirming that recovery was fully automatic across the benchmark.

\begin{table}[t]
\centering
\caption{Per-stage recovery on NopSCADlib. Trigger rate is the fraction of samples invoking a layer; recovery rate is its post-recovery success rate; retries is the mean number of attempts when invoked. The manual-repair layer is omitted: it was never invoked in these experiments.}
\label{tab:recovery}
\small
\setlength{\tabcolsep}{5pt}
\begin{tabular}{lccc}
\toprule
Layer & Trigger & Recovery & Retries \\
\midrule
schema repair          & 0.072 & 1.000 & 1.00 \\
SCAD auto-fix          & 0.007 & 1.000 & 1.00 \\
VLM self-correction    & 0.804 & 0.503 & 1.62 \\
component verification & 0.173 & 0.000\textsuperscript{*} & 1.61 \\
\bottomrule
\end{tabular}

\vspace{2pt}
{\footnotesize\raggedright \textsuperscript{*}Verification is scored by part presence, not the shared success flag (essential/secondary pass: 94.6\%/97.2\%).\par}
\end{table}

\section{Discussion}
\label{Sec:discussion}

CRAFT achieves competitive perceptual and geometric quality and much greater editability than direct one-shot generation, with the clearest gains on complex and out-of-distribution designs. This supports the view that structured generation becomes more valuable as geometric dependencies and design complexity grow. The JSON intermediate representation is designed to preserve symbolic dependencies during code generation; its ablation produces a similar Chamfer distance but lower F1 and voxel IoU, indicating that it also affects geometric structure. The editability evaluation further shows that the complete CRAFT framework exposes considerably more parameters and symbolic relationships than direct generation, although isolating the JSON IR's contribution to editability would require an editability-specific ablation.

Several \textbf{limitations} should be noted. Performance declines as complexity increases, particularly when shapes involve many interacting parts or implicit structural relationships that are difficult to infer from text alone. OpenSCAD, while well suited for symbolic parametric modeling, does not capture the feature-history richness of industrial CAD systems such as Fusion 360 Gallery~\citep{fusion360gallery2021}. The multi-stage framework also depends on external language and vision models, increasing inference cost and introducing the risk of error propagation across stages. Finally, the evaluation covers three datasets, which represents only a subset of real-world designs.

\section{Conclusions}
\label{Sec:conclusion}

We presented CRAFT, a multi-agent framework for text-to-parametric CAD generation that combines structured reasoning, a JSON-based intermediate representation, visual verification, and layered error recovery. CRAFT explicitly preserves symbolic parametric relationships during generation, producing OpenSCAD programs that are executable, editable, and reusable. Experiments across three datasets show that CRAFT achieves competitive perceptual and geometric quality and markedly higher editability, and an ablation isolates retrieval as the dominant contributor to in-domain volumetric overlap, with layered recovery contributing most to Chamfer distance and F1. These results demonstrate that explicitly preserving parametric intent is a promising and practical direction for text-to-CAD systems.

\clearpage
{
\small
\bibliographystyle{IEEEtran}
\bibliography{Refs}
}

% ============================================================
% APPENDICES — arXiv version only.
% Everything ABOVE this line is identical to the ICLAD 2026
% camera-ready source (main paper), except the \IEEEpubid line,
% which carries the IEEE accepted-manuscript notice required for
% arXiv posting. Everything BELOW is extra material that did not
% fit the conference page limit. Appendices are lettered A, B, ...;
% table and figure numbers continue from the main paper.
% ============================================================
\clearpage
\newcommand{\todo}[1]{\textcolor{red}{[TODO: #1]}}
\newcommand{\tbd}{\textcolor{red}{TBD}}
\renewcommand{\dbltopfraction}{0.9}

\section*{Supplementary Material}
The appendices below give implementation details and results omitted from the ICLAD 2026 version for space.

\appendices

\section{Additional Experimental Details}
\label{app:details}
\secref{Sec:setup} omits the following implementation details for space.

\textbf{Prompt-complexity scoring.} For NopSCADlib, complexity tiers are assigned by source-code analysis. Each component receives a weighted composite of seven structural features: primitive count ($\times1.0$), Boolean operations ($\times1.5$), CSG depth ($\times0.5$), parameter count ($\times0.3$), sub-module calls ($\times2.0$), lines of code ($\times0.05$), and transform count ($\times0.3$). The tiers are split at the 33rd and 66th percentiles of the composite score (boundaries 36.50 and 66.25), which yields 159 Simple, 160 Medium, and 149 Complex components. For ABC and Slice-100K, source code is unavailable, so tiers are assigned by mesh-geometry analysis using five weighted features: face count (0.25), connected components (0.15), convex hull ratio (0.25), surface-to-bounding-box ratio (0.20), and compactness (0.15), with the same percentile split. These weights were chosen heuristically and are intended only to group components by relative difficulty within each dataset.

\textbf{Sampling temperature.} CRAFT samples at temperature $0.0$ for compilation, matching the direct-generation baselines, at $0.1$ for planning and code repair, and at $0.2$ for understanding and for the VLM and verification assessments. \secref{Sec:method} and \secref{Sec:setup} give the model allocation and iteration budgets.

\section{Full Ablation Results}
\label{app:ablation}
\tabref{tab:ablation} reports the per-variant metrics underlying the ablation in \secref{subsec:ablation}.

\begin{table}[!b]
\centering
\caption{Ablation on NopSCADlib: full framework versus single-component-disabled variants ($-$component). The final column summarizes how the mesh-based metrics change when the component is removed.}
\label{tab:ablation}
\footnotesize
\setlength{\tabcolsep}{3pt}
\begin{tabular}{lcccl}
\toprule
Variant & CD$\downarrow$ & F1@1\%$\uparrow$ & IoU$\uparrow$ & Effect of removal \\
\midrule
CRAFT (full)      & 0.0654 & 0.2530 & 0.2356 & reference \\
$-$\,retrieval    & 0.0764 & 0.2030 & 0.1545 & worse on all three \\
$-$\,multi-view   & 0.0738 & 0.1907 & 0.2272 & worse CD and F1 \\
$-$\,recovery     & 0.0773 & 0.1878 & 0.2289 & worse CD and F1 \\
$-$\,JSON-IR      & 0.0643 & 0.2383 & 0.1999 & worse F1 and IoU \\
$-$\,verification & 0.0637 & 0.2574 & 0.2411 & slightly better \\
\bottomrule
\end{tabular}
\end{table}

\section{Complexity-Tier Breakdown}
\label{app:complexity}
\tabref{tab:abc-results} and \tabref{tab:slice-results} break the external-dataset results of \tabref{tab:external-results} down by complexity tier, and \tabref{tab:nopscad-results} gives the same breakdown for NopSCADlib. Each external tier holds about 33 shapes, so the per-tier differences on ABC and Slice-100K are subject to more sampling noise than the overall rows.

On NopSCADlib, direct generation is competitive on simple parts, many of which are single primitives well within a strong model's direct-generation distribution. On this tier, GPT-5.2 has the lowest Chamfer distance, GPT-4o has the highest voxel IoU, and CRAFT has the highest F1. GPT-5.2 leads all three metrics on the medium tier. On the complex tier, CRAFT has the lowest Chamfer distance and the highest voxel IoU but the lowest F1 of the three methods (0.149, against 0.169 for GPT-4o and 0.188 for GPT-5.2). CRAFT's complex-tier gains therefore appear in Chamfer distance and volumetric overlap rather than in surface agreement at the 1\% threshold.

On ABC, CRAFT has the lowest Chamfer distance and the highest F1 on every tier, and the highest voxel IoU on the simple and medium tiers. GPT-5.2 exceeds CRAFT on complex-tier voxel IoU by 0.0003.

On Slice-100K, the ordering is mixed. CRAFT has the highest F1 on every tier and the lowest Chamfer distance on the medium and complex tiers. GPT-5.2 has the lowest Chamfer distance and the highest voxel IoU on the simple tier, and GPT-4o has the highest voxel IoU on the medium and complex tiers. On the complex tier, CRAFT and GPT-5.2 differ by less than 0.001 in both Chamfer distance and voxel IoU.

CRAFT has the lowest complex-tier Chamfer distance on all three datasets, while its complex-tier F1 and voxel IoU rankings vary by dataset. On simple parts, the comparison also varies: CRAFT leads every metric on ABC but trails a direct baseline on Chamfer distance and voxel IoU on NopSCADlib and Slice-100K.

\begin{table}[t]
\centering
\caption{Geometric evaluation on ABC (retrieval disabled), with breakdown by complexity tier. F1 at a 1\% threshold. Best in each row group in bold.}
\label{tab:abc-results}
\small
\setlength{\tabcolsep}{6pt}
\begin{tabular}{llccc}
\toprule
Split & Method & CD$\downarrow$ & F1$\uparrow$ & Voxel IoU$\uparrow$ \\
\midrule
\multirow{3}{*}{Overall}
& CRAFT   & \textbf{0.0841} & \textbf{0.1233} & \textbf{0.0433} \\
& GPT-4o  & 0.1008 & 0.0857 & 0.0212 \\
& GPT-5.2 & 0.0969 & 0.0875 & 0.0192 \\
\midrule
\multirow{3}{*}{Simple}
& CRAFT   & \textbf{0.0804} & \textbf{0.1134} & \textbf{0.0346} \\
& GPT-4o  & 0.1133 & 0.0574 & 0.0199 \\
& GPT-5.2 & 0.1034 & 0.0665 & 0.0089 \\
\midrule
\multirow{3}{*}{Medium}
& CRAFT   & \textbf{0.0893} & \textbf{0.1212} & \textbf{0.0637} \\
& GPT-4o  & 0.0914 & 0.0856 & 0.0298 \\
& GPT-5.2 & 0.0917 & 0.1127 & 0.0150 \\
\midrule
\multirow{3}{*}{Complex}
& CRAFT   & \textbf{0.0829} & \textbf{0.1339} & 0.0330 \\
& GPT-4o  & 0.0954 & 0.1196 & 0.0143 \\
& GPT-5.2 & 0.0953 & 0.0842 & \textbf{0.0333} \\
\bottomrule
\end{tabular}
\end{table}

\begin{table}[t]
\centering
\caption{Geometric evaluation on Slice-100K (retrieval disabled), with breakdown by complexity tier. F1 at a 1\% threshold. Best in each row group in bold.}
\label{tab:slice-results}
\small
\setlength{\tabcolsep}{6pt}
\begin{tabular}{llccc}
\toprule
Split & Method & CD$\downarrow$ & F1$\uparrow$ & Voxel IoU$\uparrow$ \\
\midrule
\multirow{3}{*}{Overall}
& CRAFT   & \textbf{0.0612} & \textbf{0.2924} & 0.0616 \\
& GPT-4o  & 0.0751 & 0.2122 & \textbf{0.0711} \\
& GPT-5.2 & 0.0622 & 0.2768 & 0.0636 \\
\midrule
\multirow{3}{*}{Simple}
& CRAFT   & 0.0718 & \textbf{0.2593} & 0.0522 \\
& GPT-4o  & 0.0796 & 0.2017 & 0.0504 \\
& GPT-5.2 & \textbf{0.0655} & 0.2452 & \textbf{0.0556} \\
\midrule
\multirow{3}{*}{Medium}
& CRAFT   & \textbf{0.0553} & \textbf{0.3034} & 0.0617 \\
& GPT-4o  & 0.0715 & 0.2067 & \textbf{0.0841} \\
& GPT-5.2 & 0.0641 & 0.2787 & 0.0645 \\
\midrule
\multirow{3}{*}{Complex}
& CRAFT   & \textbf{0.0568} & \textbf{0.3139} & 0.0709 \\
& GPT-4o  & 0.0743 & 0.2282 & \textbf{0.0788} \\
& GPT-5.2 & 0.0569 & 0.3065 & 0.0707 \\
\bottomrule
\end{tabular}
\end{table}

\begin{figure}[t]
    \centering
    \includegraphics[width=0.99\linewidth,trim={1in 2.3in 1in 2.3in},clip]{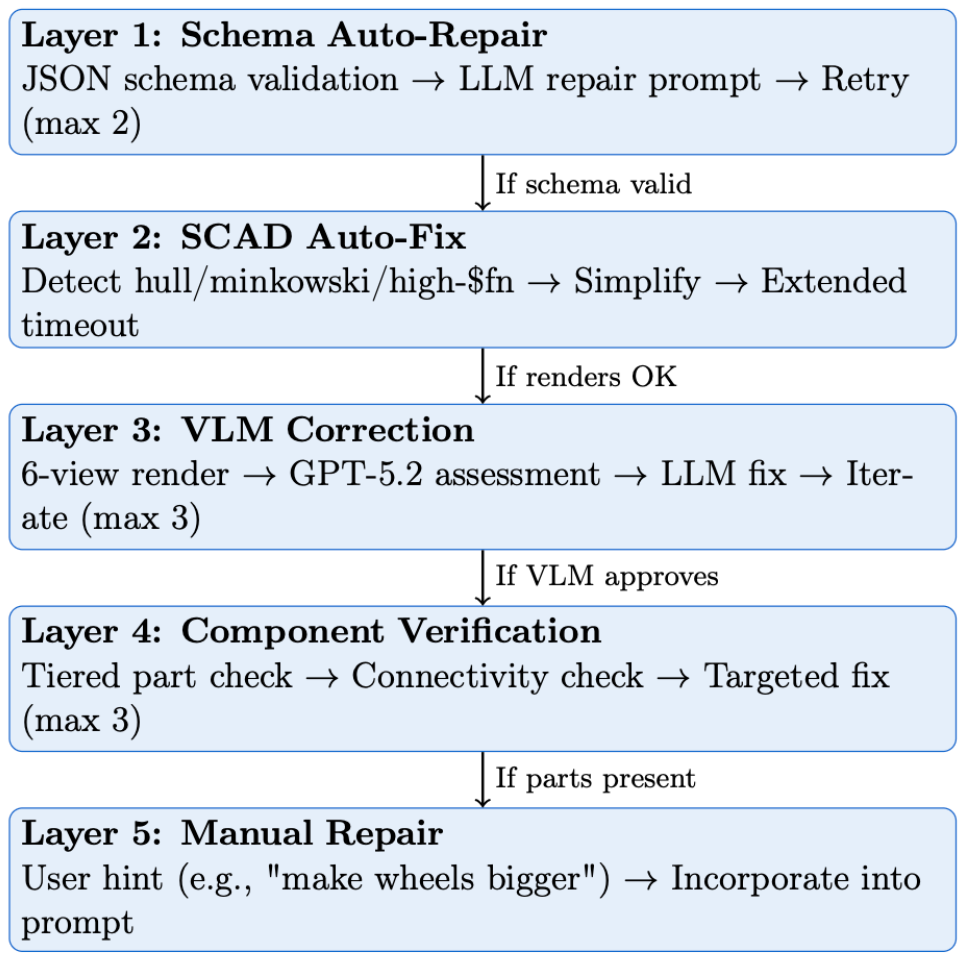}
    \caption{The five layers of CRAFT's recovery strategy: schema auto-repair, SCAD auto-fix, VLM self-correction, component verification, and optional manual repair. Repairs escalate from structural fixes to semantic correction, with bounded retry budgets at intermediate layers.}
    \label{fig:recovery-layers}
\end{figure}

\begin{figure*}[t!]
    \centering
    \includegraphics[width=0.8\textwidth]{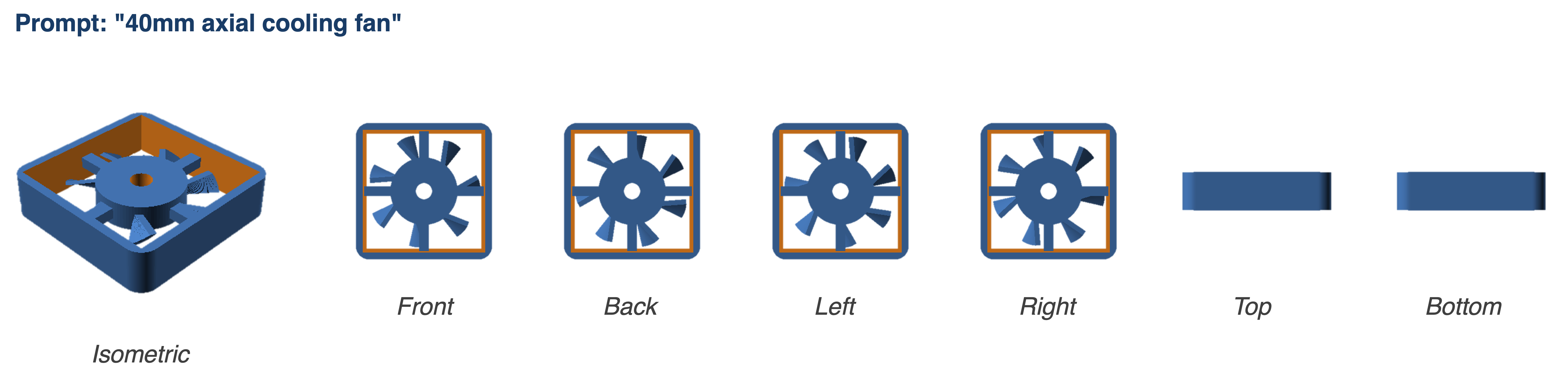}
    \includegraphics[width=0.8\textwidth]{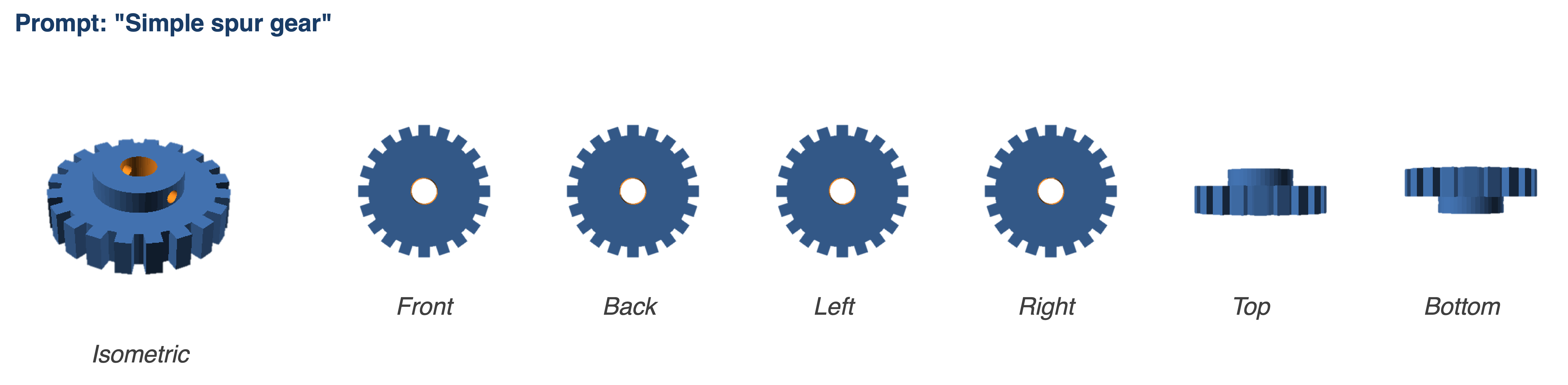}
    \includegraphics[width=0.8\textwidth]{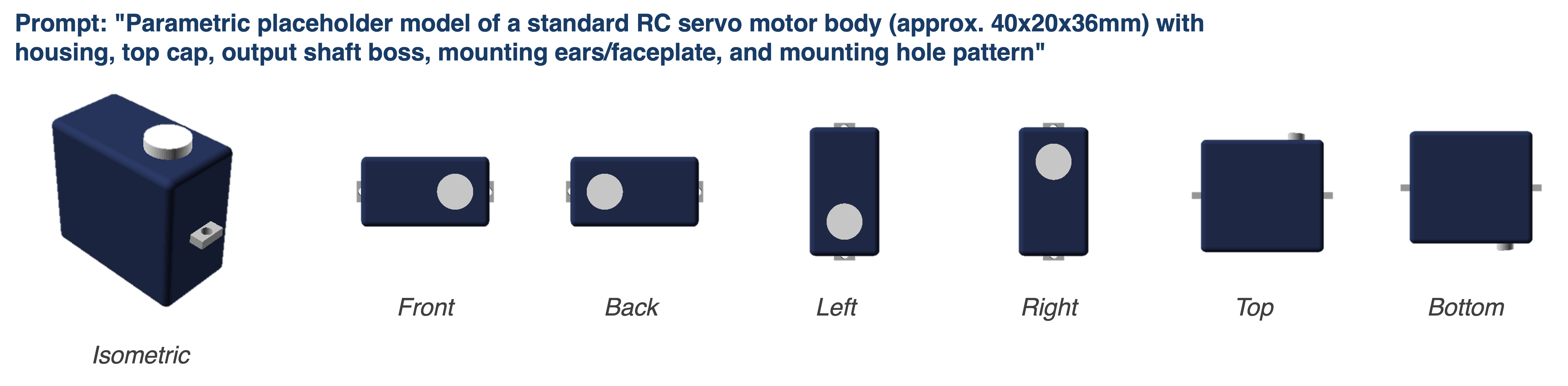}
    \includegraphics[width=0.8\textwidth]{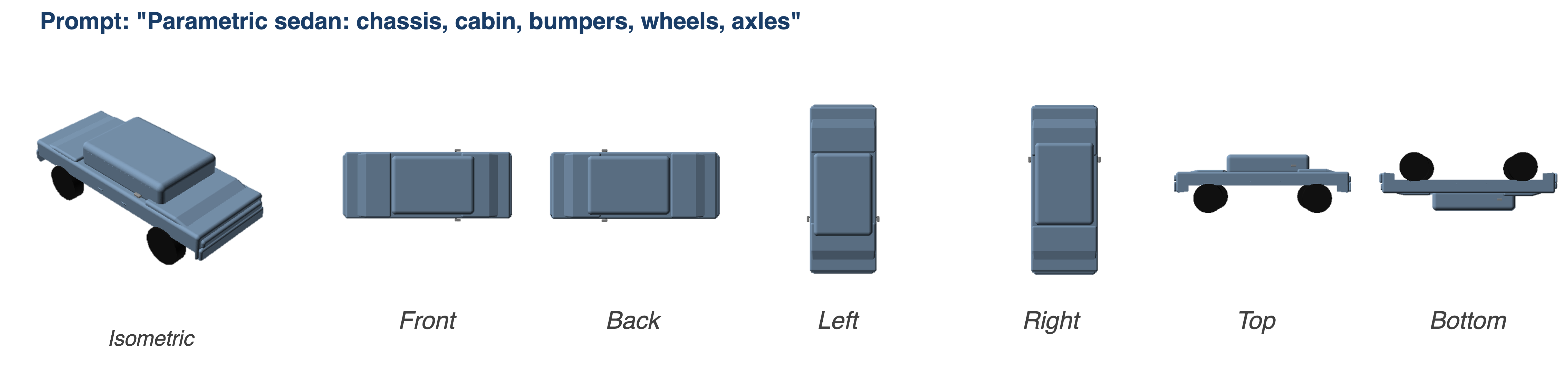}
    \includegraphics[width=0.8\textwidth]{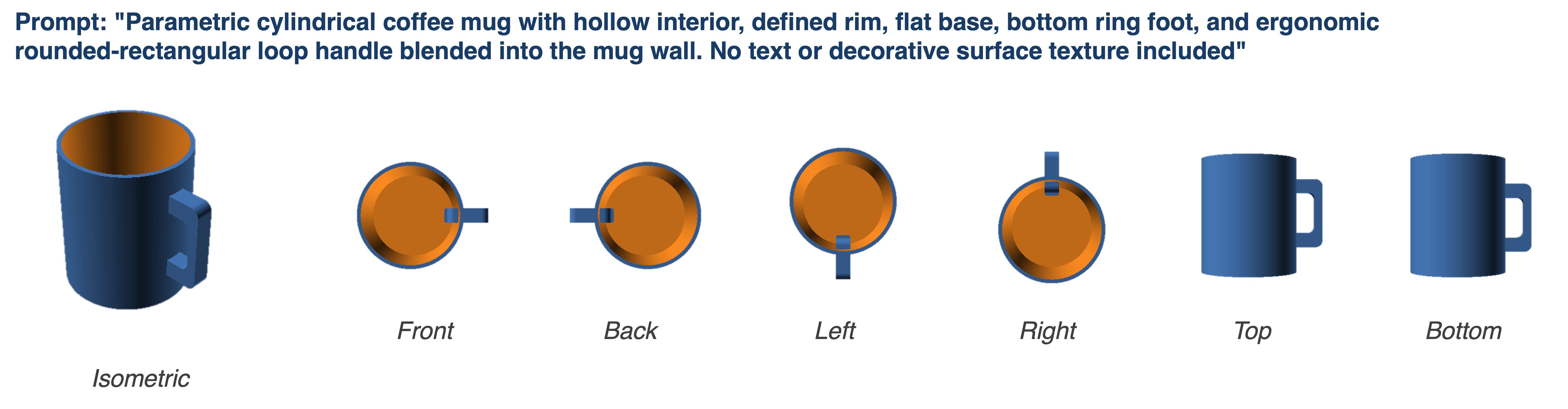}
    \includegraphics[width=0.8\textwidth]{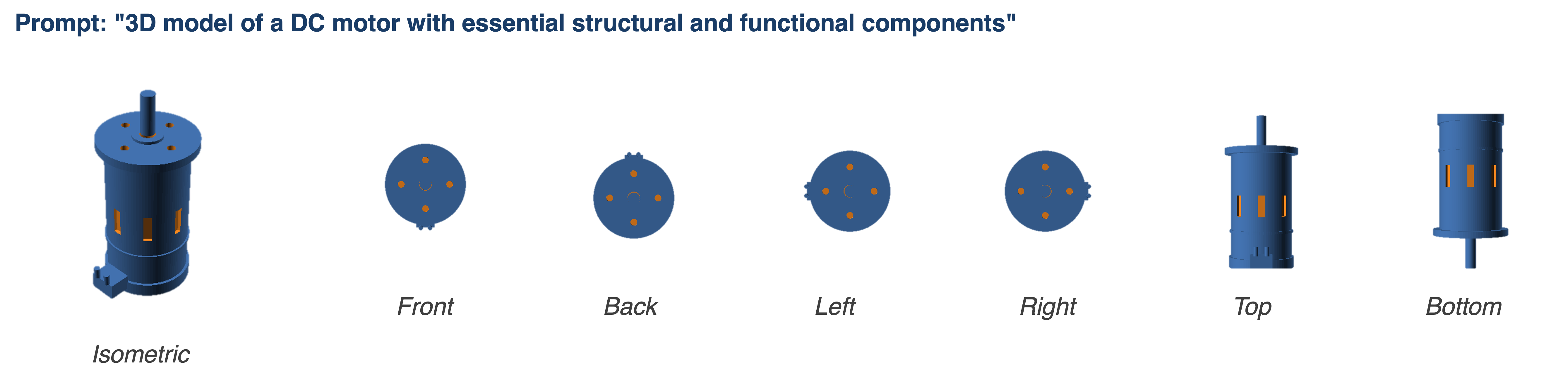}
    \caption{Multi-view rendering outputs from CRAFT. Each row shows a model generated from a text prompt: the isometric output alongside the six orthographic views used for VLM assessment.}
    \label{fig:multiview-assessment}
\end{figure*}

\section{Matched-Effort Baselines}
\label{app:matched}
To test whether CRAFT's advantage comes from issuing more model calls, we give direct generation with GPT-5.2 additional calls in three ways. Direct + repair follows one-shot generation with the SCAD auto-fix of the second recovery layer. Direct + VLM follows one-shot generation with the multi-view VLM correction of Stage~5. Direct $\times$5 invokes direct generation five times, at temperatures from 0.0 to 0.3, and keeps the candidate scored highest by CRAFT's deterministic validator, with ties going to the earliest candidate. The validator weights syntax validity (0.35), a non-blank render (0.35), non-degenerate primitive dimensions (0.15), and keyword coverage of the expected part (0.15). The five calls round up CRAFT's mean of 4.7 calls per sample under the recovery budget of \secref{subSec:recovery}, which counts the three forward-path calls and one call per retry. This count excludes the initial VLM assessment and verification round and the correction calls that follow failed checks. Counting every API call, CRAFT makes 9.5 calls per sample (8.5 GPT-5.2 and 1.0 GPT-4o), of which 3.6 are made by VLM self-correction and 2.8 by component verification. The augmented baselines therefore receive fewer calls than CRAFT (\tabref{tab:matched}). Call count is only an approximate measure of budget, because CRAFT mixes GPT-5.2 and GPT-4o calls, which differ in token cost. Token usage and API cost were not logged. Mean wall-clock time per sample, including OpenSCAD rendering, is 86~s for CRAFT, against 10~s for Direct + repair, 34~s for Direct + VLM, and 33~s for Direct $\times$5. In addition to the metrics of \secref{Sec:setup}, \tabref{tab:matched} reports F1 at a 5\% distance threshold (F1@5\%), which measures coarse shape agreement.

At matched compute, the geometric metrics largely converge. CRAFT has the lowest Chamfer distance, the highest F1@5\%, and the highest voxel IoU, and Direct $\times$5 leads F1@1\%. CRAFT's margins over Direct + repair and Direct $\times$5 are at most 0.003 in Chamfer distance and 0.014 in voxel IoU. Neither baseline adds much to a single call. All 468 Direct + repair programs rendered on the first attempt, so its repair step never ran. Direct $\times$5 kept its first candidate on 408 of the 468 prompts, because the validator rarely separates renderable candidates (all five tied on 314 prompts). Both are therefore close to a single GPT-5.2 call, and they differ from the single-call baseline of \secref{Sec:setup} mainly in prompt wording. Direct + repair nonetheless exceeds that baseline by 0.033 in voxel IoU, most of CRAFT's 0.047 gain over it. CRAFT's voxel-IoU gain over a single direct call therefore lies largely within the sensitivity of direct generation to its prompt. The comparisons in \secref{Sec:results}, including the gains on complex and out-of-distribution designs, are against single-call baselines rather than budget-matched ones. Two differences remain at equal budget: retrieval grounding, whose removal causes the largest drop in voxel IoU in the ablation (\secref{subsec:ablation}), and the number of exposed parameters examined below.

Direct + VLM is the weakest augmented baseline. Its original run exported meshes only for the 317 outputs that the VLM approved, whereas CRAFT keeps its output regardless of approval. We therefore evaluate Direct + VLM over all 468 prompts, as for the other methods, and eight of its programs fail to export. On the 458 prompts evaluated for both, Direct + VLM is worse than Direct + repair, which uses the same prompt without VLM correction, on all three metrics (Chamfer distance 0.078 against 0.069, F1@1\% 0.218 against 0.261, and voxel IoU 0.195 against 0.223). Removing VLM correction from CRAFT instead raises Chamfer distance and lowers F1 (\secref{subsec:ablation}). VLM correction therefore improves geometry when it refines CRAFT's planned programs but degrades it when applied to direct generation.

Under matched compute, CRAFT's editability advantage narrows (\tabref{tab:edit-matched}). The augmented baselines expose 8.8 or 6.0 parameters per model against CRAFT's 13.1, a factor of 1.5 to 2.2, and all three expose more than single-call GPT-5.2 (4.2). The symbolic-preservation rate is no longer a CRAFT advantage: Direct + VLM reaches 69.1\%, above CRAFT's 66.8\%. GPT-5.2 also writes more parametric OpenSCAD with different prompt wording or with VLM correction, and CRAFT's remaining editability advantage lies in the number of exposed parameters rather than in the symbolic-preservation rate. Edit success and post-edit validity are at least 98.6\% for CRAFT and the augmented baselines, over all 468 prompts.

\begin{table}[t]
\centering
\caption{Matched-effort baselines on NopSCADlib (GPT-5.2 direct generation augmented with repair, VLM correction, or repeated sampling). F1 is reported at 1\% and 5\% distance thresholds. LLM calls is the mean number of API calls per sample. Best among the matched-budget methods in bold.}
\label{tab:matched}
\footnotesize
\setlength{\tabcolsep}{3pt}
\begin{tabular}{lccccc}
\toprule
Method & CD$\downarrow$ & F1@1\%$\uparrow$ & F1@5\%$\uparrow$ & IoU$\uparrow$ & LLM calls \\
\midrule
CRAFT                          & \textbf{0.0654} & 0.2530 & \textbf{0.6260} & \textbf{0.2356} & 9.5 \\
Direct + repair                & 0.0684 & 0.2576 & 0.6066 & 0.2217 & 1.0 \\
Direct + VLM                   & 0.0783 & 0.2177 & 0.5744 & 0.1939 & 4.0 \\
Direct $\times$5               & 0.0678 & \textbf{0.2651} & 0.6111 & 0.2262 & 5.0 \\
\midrule
GPT-5.2 (single call)          & 0.0662 & 0.2587 & 0.6055 & 0.1890 & 1.0 \\
\bottomrule
\end{tabular}
\end{table}

\begin{table}[t]
\centering
\caption{Editability at matched compute on NopSCADlib. CRAFT exposes 1.5--2.2$\times$ more parameters than the augmented baselines, while the symbolic-preservation rate is comparable across methods.}
\label{tab:edit-matched}
\footnotesize
\setlength{\tabcolsep}{3pt}
\begin{tabular}{lcccc}
\toprule
Method & \# Params & Symbolic$\uparrow$ & Edit succ.$\uparrow$ & Valid$\uparrow$ \\
\midrule
CRAFT                              & \textbf{13.1} & 66.8\% & 99.9\% & 99.1\% \\
Direct + VLM                       & 8.8  & \textbf{69.1\%} & 99.4\% & 98.6\% \\
Direct + repair                    & 6.0  & 57.3\% & 100.0\% & 98.9\% \\
Direct $\times$5                   & 6.0  & 59.4\% & 100.0\% & 99.2\% \\
\midrule
GPT-5.2 (single call)              & 4.2  & 47.7\% & 99.2\% & 97.7\% \\
\bottomrule
\end{tabular}
\end{table}

\section{Layered Recovery and Multi-View Assessment}
\label{app:recovery}
\figref{fig:recovery-layers} shows the five recovery layers of \secref{subSec:recovery} with the check that gates each transition and the retry budget of each layer. A plan advances from schema repair to SCAD auto-fix once it passes schema validation, to VLM correction once it renders, and to component verification once the VLM approves it. \tabref{tab:recovery} reports how often each layer was triggered on NopSCADlib: VLM correction was triggered on 80.4\% of samples, and the two structural layers on fewer than 8\%.

\figref{fig:multiview-assessment} shows the six orthographic views passed to the VLM for six generated models. In the spur-gear row, the four face-on views show the tooth count and the bore, while only the top and bottom views show that the hub protrudes from one face of the gear. An error in hub height would therefore be invisible in four of the six views.

\end{document}